\documentclass[conference]{IEEEtran}
\IEEEoverridecommandlockouts
\usepackage{changes}
\usepackage{cite}
\usepackage{amsmath,amssymb,amsfonts}
\usepackage{algorithmic}
\usepackage{graphicx}
\usepackage{textcomp}
\usepackage{stfloats}
\usepackage{float}
\usepackage{xcolor}
\usepackage{booktabs}
\usepackage{balance}
\usepackage{siunitx}       
\newcommand{\sS}{\mathcal{S}}
\newcommand{\sA}{\mathcal{A}}
\newcommand{\sP}{\mathcal{P}}
\DeclareMathOperator*{\argmax}{arg\,max}
\def\BibTeX{{\rm B\kern-.05em{\sc i\kern-.025em b}\kern-.08em
    T\kern-.1667em\lower.7ex\hbox{E}\kern-.125emX}}
\begin{document}

\begin{figure*}[htbp]
\centerline{\includegraphics[width=6in]{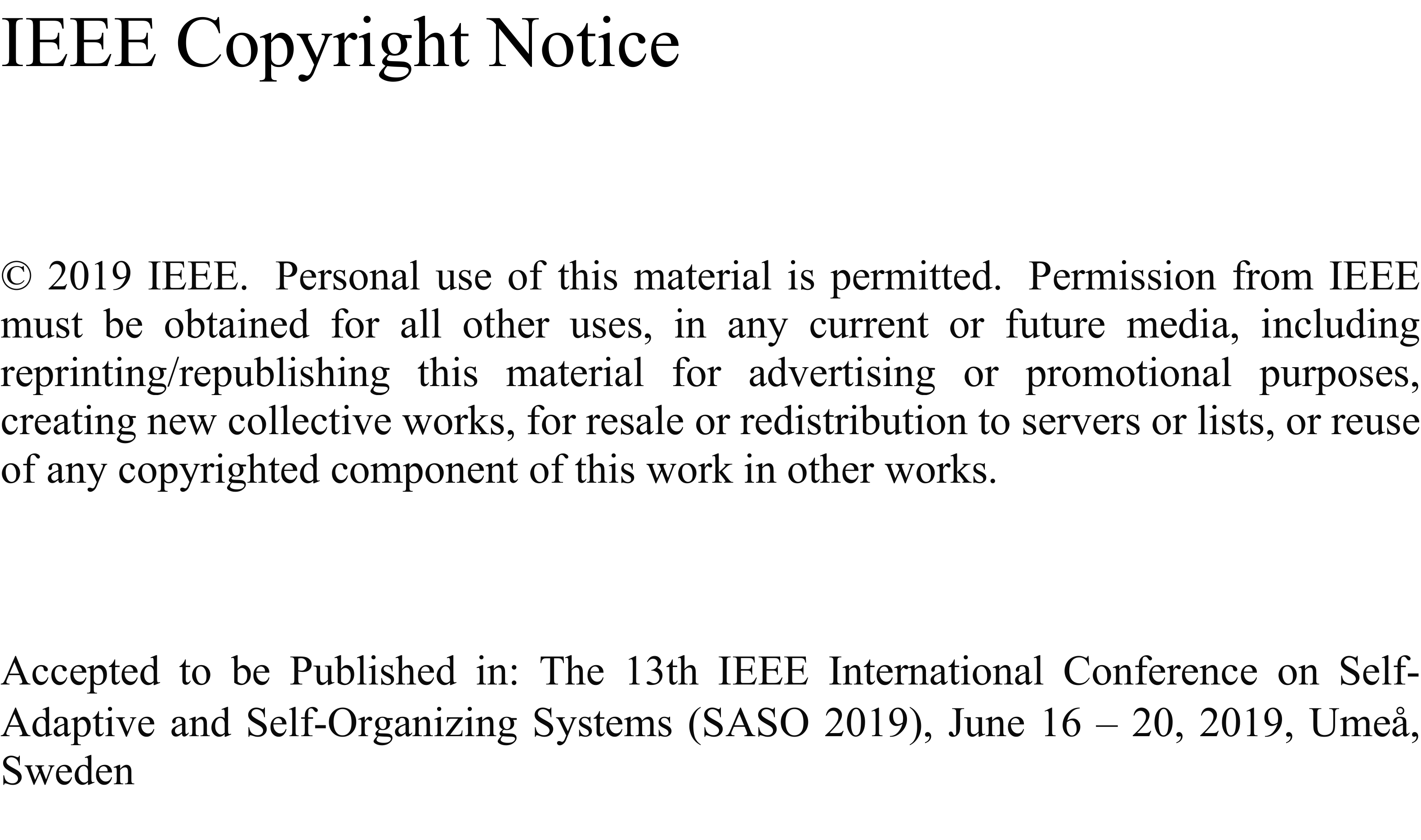}}
\end{figure*}

\title{Autonomous Management of Energy-Harvesting IoT Nodes Using Deep Reinforcement Learning}

\author{Abdulmajid Murad\textsuperscript{1}*,  Frank Alexander Kraemer\textsuperscript{1}, Kerstin Bach\textsuperscript{2}, Gavin Taylor\textsuperscript{3}\\
\textsuperscript{1}Department of Information Security and Communication Technology\\
\textsuperscript{2}Department of Computer Science\\
Norwegian University of Science and Technology, Trondheim, Norway\\
\textsuperscript{3}Department of Computer Science, United States Naval Academy, Annapolis, USA \\
*Corresponding author: abdulmajid.a.murad@ntnu.no
}
\IEEEoverridecommandlockouts
\IEEEpubid{\makebox[\columnwidth]{\copyright2019 IEEE \hfill} \hspace{\columnsep}\makebox[\columnwidth]{ }}

\maketitle
\IEEEpubidadjcol

\begin{abstract}
  Reinforcement learning (RL) is capable of managing wireless,
  energy-harvesting IoT nodes by solving the problem of autonomous management
  in non-stationary, resource-constrained settings.  We show that the
  state-of-the-art policy-gradient approaches to RL are appropriate for the
  IoT domain and that they outperform  previous approaches.  Due to the
  ability to model continuous observation and action spaces, as well as
  improved function approximation capability, the new approaches are able to
  solve harder problems, permitting reward functions that are better aligned
  with the actual application goals.  We show such a reward function and use
  policy-gradient approaches to learn capable policies, leading to behavior
  more appropriate for IoT nodes with less manual design effort, increasing
  the level of autonomy in IoT.

\end{abstract}


\section{Introduction}

Digitizing domains like smart cities, precision agriculture, manufacturing,
and healthcare requires the instrumentation of the physical world with
wireless IoT nodes to sense environmental variables.  The actual logic of an
IoT device can be rather simple, as they usually make measurements, prepare
the data and send it into a central fusion center for aggregation.  However,
challenges arise from the limited energy resources of the wireless and
miniaturized nodes.  Therefore, the choice of parameters like duty cycle (percentage of time a node is active) and
sampling frequency is crucial since they have a dominant effect on the energy
consumption and performance of the node.  Currently, such parameters are often
assigned manually to an entire family of IoT devices, and stay constant during
deployment, or are determined by manually fine-tuned algorithms.  However, IoT
in its full scale presents a setting where nodes are placed in dynamic and
non-stationary environments, so that a one-fits-all approach and manual design
efforts are impossible.  Future solutions should instead be based on
autonomous nodes that optimize their behavior by learning within their
specific environment.

Our vision for a more autonomous and hence scalable IoT is inspired by recent developments in deep reinforcement learning (RL), and its success in building capable autonomous agents with little manual design effort in complex domains.
Instead of manually adapting IoT nodes, we want engineers to only formulate the goal in the form of a reward function for the IoT nodes as close to the actual application goal as possible.

Reinforcement learning has indeed been used to control IoT nodes by various
works, especially for power management~\cite{Fraternali2018, Shresthamali2017,
Hsu2014}.  However, these approaches use older RL techniques like SARSA and
are constrained by coarsely discretized action and observation spaces, and
tackle only simpler problems.  Many approaches \cite{Liu2011, Hsu2015,
Aoudia2018} thus formulate reward functions in terms of indirect indicators like energy budget.
Those are often easier to learn, but
are imperfectly correlated with actual goals.  This not only adds manual
design effort, but also doesn't live up to the potential of RL.  Instead, RL
should be used to sort out the complicated and node-specific technicalities on
its own through learning.

In this paper, we explore deep RL as an end-to-end approach to enable autonomy within IoT nodes.
For the training, we use a policy gradient method, namely Proximal Policy Optimization (PPO,~\cite{Schulman2017}).  
We use neural networks as function approximators to learn state abstractions and effective policies directly from continuous input data. 
We also develop a reward function that closely reflects application goals. 
We automated the search for the best reward signal by defining it as hyperparameters from the application perspective.  
Additionally, we conduct a series of simulations to explore the influence of reward function design on a node's behavior.

To manage the computational complexity, we propose training the agent on a server as a part of device management~\cite{Braten2018} and update the node regularly based on a static or dynamic update interval~\cite{Fraternali2018}.
To this end, we built a sensor simulator, called Sensor Gym (based on OpenAI Gym~\cite{Brockman2016}), 
as a toolkit for training and comparing various RL algorithms in an IoT context.

The results in this paper show that modern and more successful deep RL techniques outperform the older ones, and that they make it possible to use more sensible reward functions that are based on IoT application goals.
This leads to tractable learning, less manual design efforts, and hence scalable autonomy for IoT systems.

The rest of this paper is organized as follows:
Sect.~II provides a brief overview of deep reinforcement learning,
and Sect.~III presents related work that adopted RL to optimize energy-harvesting sensor nodes.
In Sect.~IV, we describe the proposed system setup, and in Sect.~V we present a baseline for performance and suitability of PPO in an IoT by comparing it to an older RL technique and study the performance with a reward function based on energy neutrality. We then introduce a reward function closer to the application goals in Sect.~VI, and present the results in Sect.~VII.

\section{Deep Reinforcement Learning}

In reinforcement learning an agent learns to better perform a task by learning from its experiences in interacting with that task.  Mathematically, the task is expressed with a Markov Decision Process (MDP).  An MDP is a tuple $M=(\sS,\sA,\sP,R,\gamma)$, where $\sS$ is the set of all states the problem might be in; $\sA$ is the set of all actions the agent might take in response; $\sP$ is a transition kernel describing the probabilities of transitioning between two states when performing an action ($p(s'|s,a)~ \forall s,s'\in\sS,a\in\sA$); $R:\sS\rightarrow\Re$ is a reward function (for example, positive numbers in states where the agent has achieved its goal, and negative numbers in states where the agent has harmed itself); and $\gamma\in (0,1)$ is a discount factor, which is used to describe how much the agent prefers rewards in the short term to rewards in the long term.  The behavior of the agent is expressed as a policy $\pi:\sS\rightarrow\sA$.

The quality of a policy can be measured using the value function of a state $V^\pi(s)$, or, the Q-value of a state-action pair $Q^\pi(s,a)$:
\begin{align}
  \label{eqn:value}
  V^\pi(s)=&R(s)+\int_\sS p(s'|s,\pi(s))V^\pi(s')ds'\\
  Q^\pi(s,a)=&R(s)+\int_\sS p(s'|s,a)Q^\pi(s',\pi(s'))ds'
\end{align}
The goal of the agent is to learn a policy which produces high value for all states by collecting trajectory samples of the form $(s_t,a_t,r_t,s_{t+1})$, where $t$ denotes the timestep, and $s_t,s_{t+1}\in\sS,a_t\in\sA,r_t=R(s_t)$.

A variety of learning algorithms for RL exist.  One traditional approach is to begin by learning Q-functions as accurately as possible using function approximation techniques; some are used in previous work applying RL to IoT, and will be introduced here.

Samples are themselves collected using a policy $\pi_{samp}$, where $a_t=\pi_{samp}(s_t)$.  The Q-functions of this sampling policy can be learned using an algorithm called SARSA; this is known as \emph{on-policy} learning.  Alternatively, the Q-functions of an improved policy may be learned using an algorithm known as Q-learning; this is known as \emph{off-policy} learning.  In either case, given Q-function approximations, a new and likely-improved policy $\pi^*$ is implied by $\pi^*(s)=\argmax_{a\in\sA}Q^\pi(s,a)$.  With sufficient sampling, alterations to allow for exploration of the full state space, and (in the on-policy case) alternation between sampling, Q-function approximation and policy improvement steps, these approaches are proven to converge to excellent policies.  This is a convenient guarantee, but requires the state and action spaces to be finite and small enough to be easily tabulated.  More complex approaches to value learning exist which allow state and action spaces to be continuous, but this results in a loss of these guarantees, and potentially much less capable policies.  In recent years, these approaches have been further improved upon by approximating functions with deep neural networks~\cite{mnih2015human}, but they remain fundamentally difficult to apply successfully to continuous and complex domains.

In contrast with value-based methods, policy-search methods directly search for a parameterized policy $\pi _{\theta}$. The result is a stochastic policy that learns a direct mapping from states to a probability distribution over actions, where preferred actions have a higher probability of being sampled. This output of a distribution makes it suitable for continuous domain tasks.

Policy gradient methods are the most prominent of policy-search methods. They
optimize a policy by maximizing the value of sampled states
$V^{\pi_\theta}(s_t)$ by performing gradient ascent over the parameters
$\theta$.  Various approximations to this basic approach exist, with
advantages in training time and sample efficiency.  In this paper, we use
Proximal Policy Optimization (PPO), one of the most successful of these
approaches\cite{Schulman2017}. PPO is a trust region method, which seeks to
iteratively maximize performance of $\pi_\theta$ compared to its previous
iteration, without changing it too much.  This allows for continuous, stable
improvement, even in continuous and complex domains such as robotic running
and difficult Atari games.  This success in demanding domains make RL ready
for application to real-world problems such as IoT.

\section{Related Work}

Several works have already adopted RL within various contexts of IoT. An
example is a dynamic power manager for energy-harvesting
wireless sensor networks~\cite{Hsu2009a}. Here Hsu et al. used
the Q-learning algorithm to train an agent to choose an action from four
levels of duty cycles. Their environment state space is based on the distance
from energy neutrality (i.e., the difference between harvested and consumed
energy), the harvested energy, and the current battery level, while the reward
function is based on the distance from energy neutrality and the current energy
storage level. They extended this work in \cite{Hsu2009b} to include quality
of service (QoS), both in the state space and the reward function.
Additionally, in \cite{Hsu2014}, they included meeting a requested throughput
along with the energy level in the reward function and added penalizing terms
for overcharging, deep-discharge, and depletion of the energy storage.
Furthermore, in \cite{Liu2011} and \cite{Hsu2015}, they used fuzzy decision
processes to model an energy harvesting node as a fuzzy environment and used a
modified Q-learning with fuzzy reward to train an RL agent.

Since the reward function formalizes the goal of an RL setup, Rioual et al.
\cite{Rioual2018} discussed the choice of the reward function in the
management of energy-harvesting IoT nodes but covered a limited range of
design choices.  To avoid intractable learning in old RL approaches when using
an application-level reward, most of the previous work used reward shaping,
which is an alternative method to guide the learning process by rewarding the
agent for achieving subgoals or developing an approximation to the desired
behaviors.  Unless the reward shaping function is based on a state-dependent
potential, it may lead to learning suboptimal or locally-optimal policies
\cite{ng1999policy}. Almost all previous work of using RL in IoT uses
manually-designed shaped functions that produce arguably acceptable results
without justifying how well the reward frames the application goal.

RL has also been used for power allocation in energy-harvesting communication
systems. Ortiz et al. \cite{Ortiz2017} used the SARSA algorithm with linear
function approximation to learn a power allocation policy in two-hop
communication. The objective of their policy is to maximize throughput of a
communication system, but they designed a reward function based on the
total power assigned to transmissions, which they claimed was correlated. They
simulated a communication environment and approximated the state space with
discrete features that indicate battery level and its constraints, harvested
energy over an hour, characteristics of the communication link, data arrival
process, and the data buffers at the communicating nodes.  Similarly, Aoudia
et al.  \cite{Aoudia2018}, used an actor-critic method with linear function
approximation to learn approximation for both policy and value function. They
used a Gaussian policy to generate continuous values of bounded packet rates
and summarized the state space by continuous values of the current residual
energy. The objective is to maximize transmitted packet rate while sustaining
perpetual operation; hence they designed a reward function to be a
multiplication of normalized residual energy and the packet rate.

Shresthamali et al.~\cite{Shresthamali2017} used a SARSA($\lambda$) RL algorithm to develop adaptive power management for a solar-energy harvesting sensor node. To simulate a sensor node, they used a scaled-up version of a real sensor powered by a battery and a solar panel, and used solar radiation data to calculate hourly harvested energy. They designed a reward function based the distance from energy neutrality, defined as the difference between the current level of energy and the optimum battery level,  which they calculated to be $\SI{60}{\percent}$ of the battery's maximum level. They trained an agent in episodes of 24 hours and rewarded it at the end of a training episode. The state space consisted of discretized information about the battery level, the distance from energy neutrality, the harvested energy, and the weather forecast, which they approximate by calculating the total amount of energy harvested in a particular day.

Fraternali et al.~\cite{Fraternali2018} focused on the configuration of the
sampling rate of indoor energy harvesting sensors. The objective of their
agent is to maximize the number of samples while avoiding power failure, by
designing a reward function that depends on the sampling rate and a penalty
for power failure. They used Q-learning with a state space that included light
intensity, the voltage level of the energy storage, and the current sampling
rate. Another example of using RL for adaptive sampling can be found in
\cite{dias2016}. Here Dias et al. proposed using Q-learning for adaptive
sampling rate adaption to avoid oversampling, while not missing environmental
changes. They introduced a variable that describes the quality of measurement
as the difference between two consecutive measurements and, depending on a
specific application, this difference should be less than a threshold value.
The action space is a range of possible sampling intervals, while the reward
function is based on the transmission avoided and the quality of measurements.

\section{System Setup}
\begin{figure}[t]
\centerline{\includegraphics[width=\linewidth]{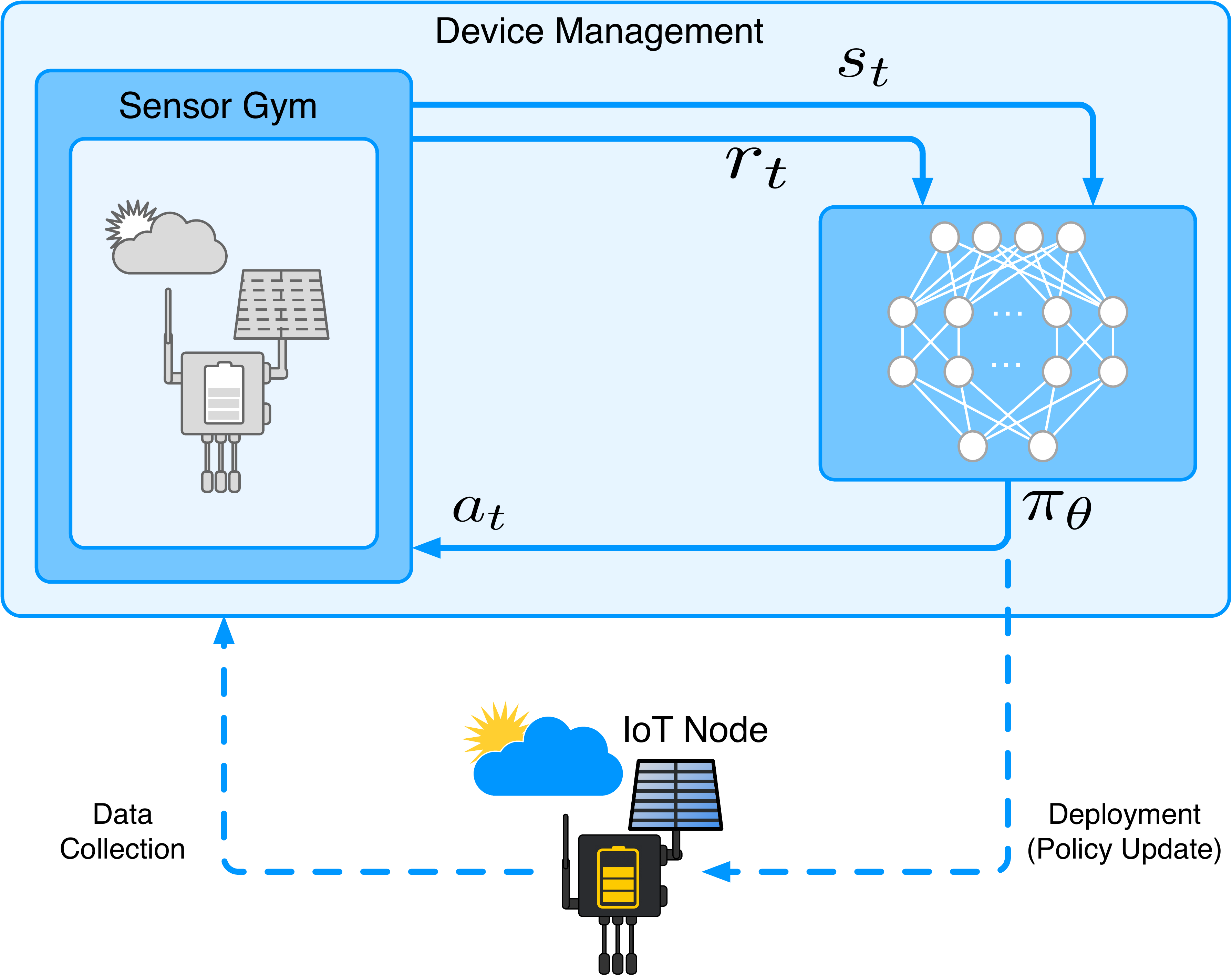}}
\caption{System setup of agent-environment interaction}
\label{sensor-gym}
\end{figure}

In this paper, we propose and apply deep RL solutions for training RL agents
to control IoT nodes autonomously.  We use PPO as a policy gradient method for
optimizing the agent's policy, using neural networks as function
approximators.  Executing the agent's policy corresponds to inference in a
neural network which has become feasible even for computationally- and
energy-constrained IoT nodes~\cite{utensor}.  We propose to integrate the much
more computation-intensive training phase on a server, as a part of the IoT
device-management~\cite{Braten2018}, illustrated in Fig.~\ref{sensor-gym}.
The deployed agents can be updated regularly based on a static or dynamic
update interval~\cite{Fraternali2018}.

To train the agent, we built a simulator of a general sensor node with an energy-harvesting solar panel, an ideal energy buffer, and a load with variable duty-cycles. We base the simulator on the OpenAI Gym, which is a toolkit for developing and comparing RL algorithms~\cite{Brockman2016}. 
We base our sensor specification on a realistic energy-harvesting IoT node. The simulated sensor has an energy buffer with a maximum capacity of $B_{\mathit{max}}= \SI{40}{\watt}$ and has a load that consumes energy during a time step according to its duty cycle.  The maximum consumed energy in an hour is  $Ec_{\mathit{max}}= \SI{0.5}{\watt\hour}$, which corresponds to a duty cycle of $D_{\mathit{max}} = \SI{100}{\percent}$ during that hour. We simulate the harvested energy by calculating the energy generated by a $\SI{6}{\watt}$ solar panel using solar radiation data.
At time step $t$, we represent the level of the energy buffer as $B_t$, the harvested energy as $\mathit{Eh_t}$, the consumed energy as $Ec_t$, and the duty cycle, which is determined by the node's policy, as $D_t$.
The consumed energy is calculated according to:

\begin{equation}
\mathit{Ec_t}= \left\{ \begin{array}{lcl}
  5 \mathit{D_t}& \mbox{if} & B_t> 5D_t \\
 0 & \mbox{if} & B_t\leq 5 D_t
\end{array}\right.
\end{equation}
The next level of the energy buffer is calculated according to:

\begin{equation}
B_{t+1}=\min(B_t+Eh_t-Ec_t,B_{max}).
\end{equation}

To handle a continuous action space $\sA=[0,D_{max}]$, the policy resulting from PPO is defined as a Gaussian distribution with mean $\mu(s_t)$ and standard deviation $\sigma(s_t)$, both of which are approximated by a neural network with parameters $\theta$:
\begin{equation}
 \pi_{\theta}(\mathit{a_t|s_t}) = \frac{\displaystyle1}{\displaystyle\sigma(\mathit{s_t})\sqrt{2\pi}}exp\left(-\frac{\displaystyle(\mathit{a_t-\mu(s_t)})^2}{\displaystyle2\sigma(\mathit{s_t})^2}\right) 
 \label{gaussian_policy}
\end{equation}
Thus, the actions are real numbers, chosen from the normal distribution in \ref{gaussian_policy}. Our neural network architecture has an output layer with two neurons which are the approximation of the mean and standard deviation of the Gaussian policy.

In Section \ref{sec:theirReward}, we demonstrate that PPO outperforms older RL approaches on previously-proposed reward functions.  Then, in Section \ref{sec:ourReward}, we demonstrate that PPO can develop good policies for reward functions that make the domain more difficult, but better capture desired behavior.  We also illustrate that older RL approaches cannot produce similarly competent policies.

\section{Reward Function Based on Energy Neutrality}
\label{sec:theirReward}

To get a baseline for the performance and suitability, we first apply PPO in the same setting as Shresthamali et al.~\cite{Shresthamali2017}, who used the SARSA($\lambda$) algorithm and designed a reward function based on the distance from energy neutrality. The concept of energy-neutrality was introduced by Kansal et al.~\cite{Kansal2007}, which states that a node is in energy-neutral operation if the consumed energy is less than or equal to the harvested energy. Accordingly, the distance from energy neutrality is redefined as the difference between the current level of the energy buffer $B_t$ and the optimum buffer level $B_0$ to account for the variance in the harvested energy over a period:
\begin{equation}
\mathit{Edist_t}= |B_t - B_0|.
\end{equation}
In~\cite{Shresthamali2017}, the reward function is formulated in terms of this distance to energy neutrality:

\begin{equation}
R_E(s_t) = \left\{ \begin{array}{lcl}
500 & \mbox{if} & \mathit{Edist_t}  = \SI{0}{\watt\hour} \\
500 - \frac{\mathit{Edist_t} } {10}& \mbox{if} & \SI{0}{\watt\hour} <\mathit{Edist_t}  \leq \SI{1}{\watt\hour}\\
10- \frac{\mathit{Edist_t}}{100} & \mbox{if} & \SI{1}{\watt\hour}< \mathit{Edist_t}  \leq \SI{5}{\watt\hour} \\
-500 & \mbox{if} & \SI{5}{\watt\hour} < \mathit{Edist_t}  
\end{array}\right.
\label{r_e}
\end{equation}
The action space is defined as discrete values of five duty cycles, which are 
\begin{equation}
a_t\in\sA= \{\SI{20}{\percent}, \SI{40}{\percent}, \SI{60}{\percent}, \SI{80}{\percent}, \SI{100}{\percent}\}
\end{equation}

The observation space contains vectors with four observations:
\begin{equation}
s_t= [B_t, \mathit{Edist_t}, \mathit{Eh_t}, \mathit{Wf_{day}}],
\end{equation}
where $\mathit{Wf_{day}}$ represents the weather forecast of the day. In \cite{Shresthamali2017}, actual harvested energy data in a day are calculated before the training and used to give information about the expected weather. The agents can leverage this information to plan their energy expenditure accordingly.

While the state space requires discretization for SARSA with a hand-designed mapping, policy approximation has the advantage of handling continuous state spaces by generalization, hence eliminating another step of manual mapping an IoT problem to RL.

\begin{figure}[t]
\centerline{\includegraphics[width=\linewidth]{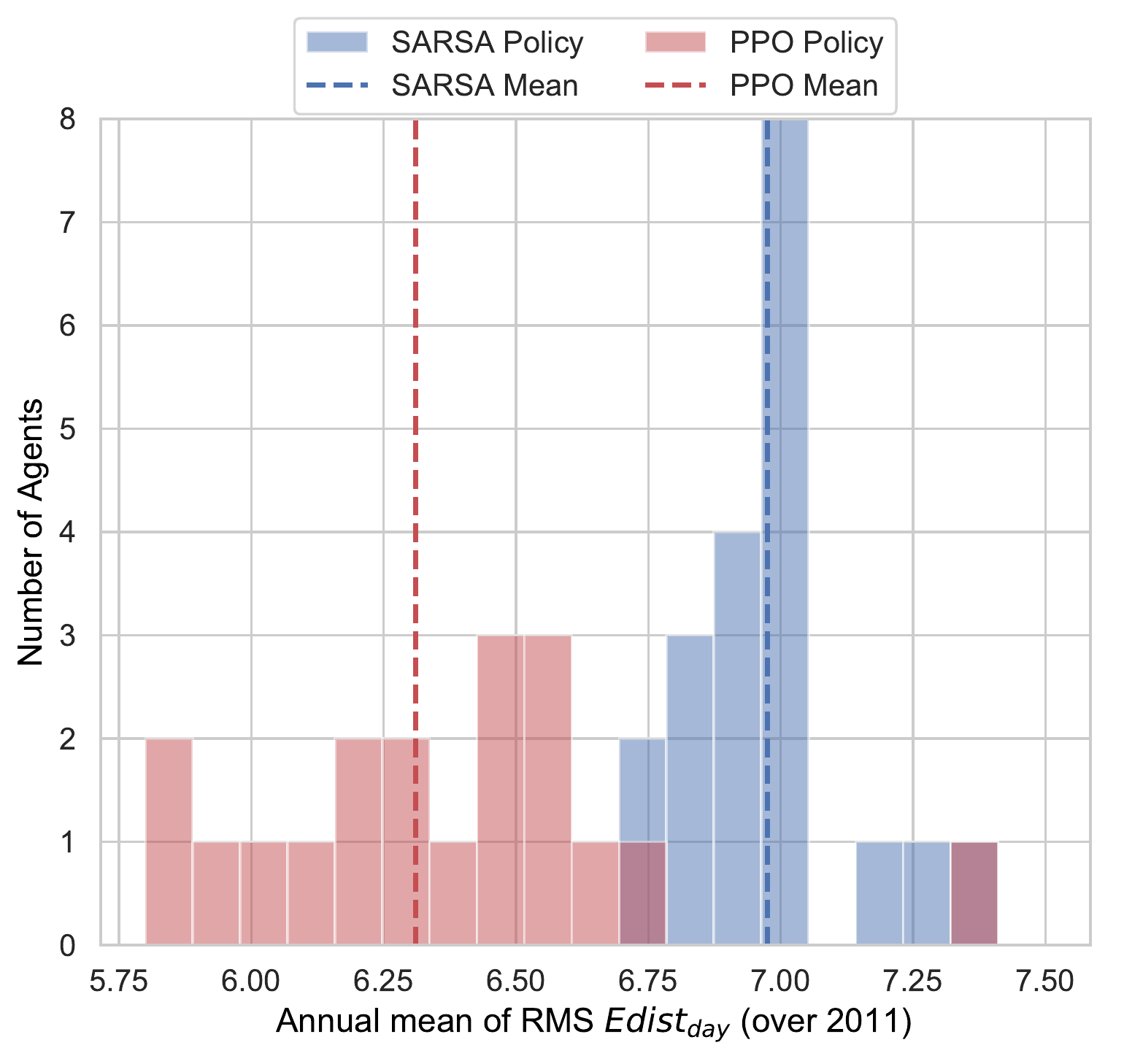}}
\caption{Performance comparison of 20 agents trained with PPO and SARSA($\lambda$)}
\label{histogram}
\end{figure}

To compare the capabilities of SARSA and PPO, we trained 20 agents with PPO and 20 with SARSA with data of Tokyo 2010 and evaluated their policies on data of Tokyo 2011, using the same settings and data as~\cite{Shresthamali2017}.
Figure~\ref{histogram} represents a histogram for performance results of both policies. 
Here, the root mean square (RMS) of $\mathit{Edist_{day}}$ values are used to measure the deviation from energy-neutrality after a one-day window and the mean of daily deviation over the whole year to compare policies. We observe that the average PPO agent is considerably better than the best SARSA policy and that almost all PPO agents are better than the average SARSA agent. 
These excellent results support our motivation to use PPO for IoT agents.

\begin{figure*}[htbp]
\centerline{\includegraphics[width=\textwidth]{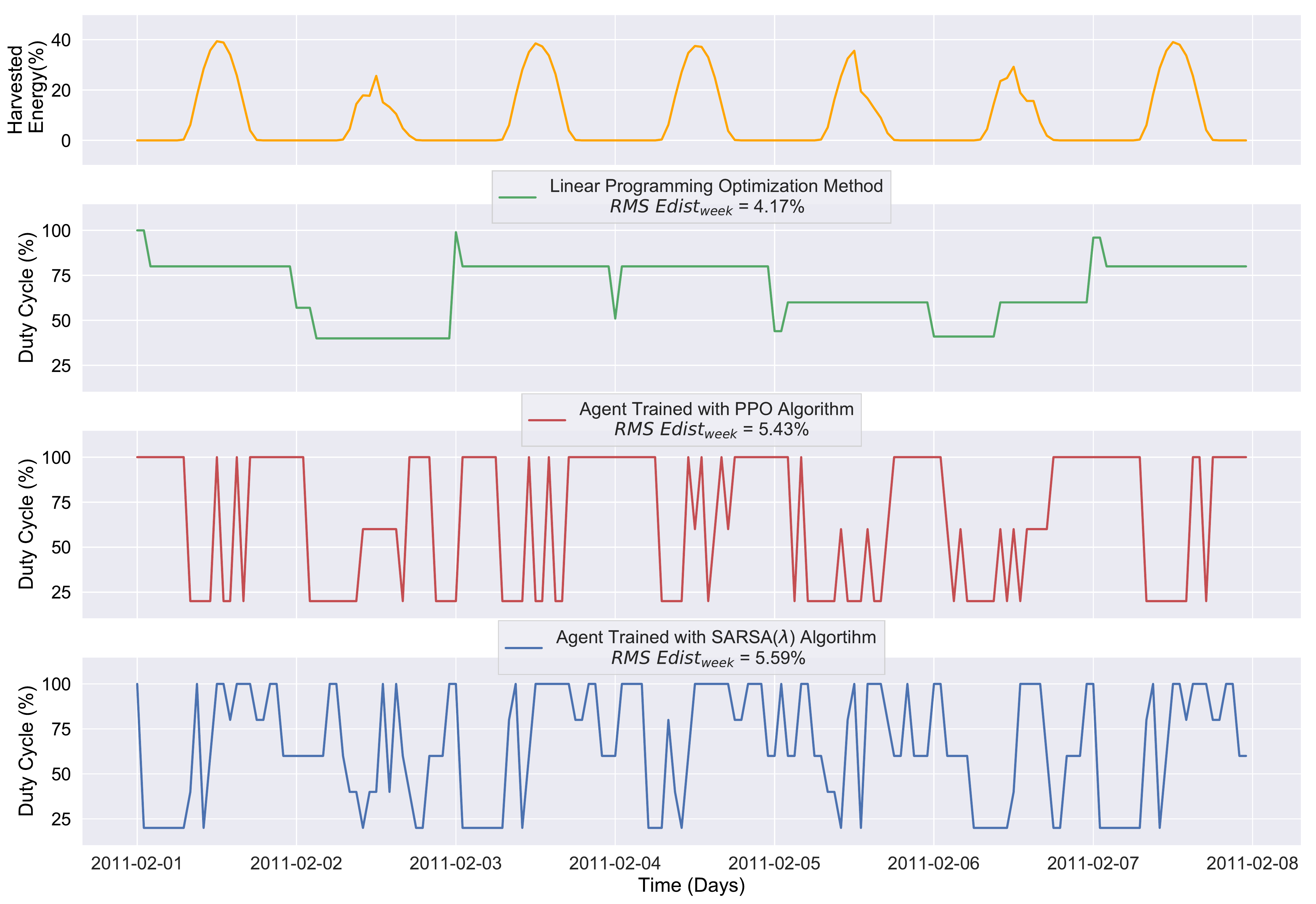}}
\caption{Comparison of PPO policy to SARSA($\lambda$) and offline policy for a week of Tokyo 2011}
\label{Tokyo_spring}
\end{figure*}

We also take a more detailed look at the behavior of the agents by comparing their results over a one-week window. Figure~\ref{Tokyo_spring} shows the performance results of PPO, SARSA policies, and linear programming when running them over one week of Tokyo weather data in February of 2011. The linear programming policies are presented in \cite{Kansal2007}, and use linear programming and acausal data (i.e., the exact future energy intake, hence not practical) to determine the optimal duty cycle given the energy neutrality constraint; 
this represents the upper limit of performance. The three policies started the week off with an energy level of $B_t = \SI{60}{\percent}$ of $B_{\mathit{max}}$ and try to maintain them at this level. RMS  $\mathit{Edist_{day}}$ values are used to measure the deviation from energy-neutrality over the whole week, and we observe that the SARSA-trained policies have the highest deviation of ($\SI{5.59}{\percent}$), PPO-trained policies have ($\SI{5.43}{\percent}$)  and the optimal policy based on linear programming has ($\SI{4.17}{\percent}$).

Figure~\ref{Tokyo_spring} illustrates a problem with the reward function based
on distance from energy neutrality (\ref{r_e}): The duty cycles of both RL
policies are subject to high variance, often oscillating between highest and
lowest duty cycle, instead of choosing a smoother course with more mid-range
values.  From an RL point-of-view, this maximizes the reward, but it is not a
behavior appropriate for IoT nodes.  Typically, we want to cover a phenomenon
as continuously as possible, and spread out measurements over time, which
corresponds to a smoother duty cycle.

Because PPO is so capable on this easier problem, in the next section we design a new, more difficult reward function, which no longer rewards the intermediate goal of energy neutrality, but more clearly expresses the goals for the agent.

\section{Reward Function Based on Application Goals}
\label{sec:ourReward}

Our objective in this section is to create a reward function that is closer to the actual application goals of an IoT system in order to encourage more desirable behavior. In particular, we want to depart from the inclusion of energy terms in the reward function, since energy is only a resource to be managed, but not a goal to be optimized.
Instead, we want the agent to maximize the sum of the duty cycle over time $\Gamma$: 
\begin{equation}
G_{\mathit{D}} =  \sum\limits_{t=0}^{\Gamma} D_t.
\label{G1}
\end{equation}
At the same time, IoT nodes must not empty their energy buffer completely.
Not only would this prevent the node from taking any measurements until enough
energy is harvested again, but it could also lead to the loss of data or leave
the node in an undefined state.  Since the sensor gym simulates a failure when
the idealized buffer is emptied, we want to minimize the occurrence of
failures over time $\Gamma$
\begin{equation}
G_{\mathit{\digamma}} = \sum\limits_{t=0}^{\Gamma} \left\{ \begin{array}{lcl}
1 & \mbox{if} & B_t=0  \\
0  & \mbox{if} & B_t>0
\end{array}\right.
\label{G2}
\end{equation}
To ensure a smooth course of the duty cycle, corresponding to a continuous stream of measurements, we want to minimize the variance of the selected duty cycles, and so minimize
\begin{equation}
G_{\mathit{Var}} = \sum\limits_{t=0}^{\Gamma} \mathit{Var_t},
\label{G3}
\end{equation}
where the variance  in the duty cycle is defined in an epoch $t$ as the absolute difference to the previous duty cycle
\begin{equation}
\mathit{Var}_t =|D_t - D_{t-1}|.
\label{var}
\end{equation}

We combine the conflicting objectives (\ref{G1}), (\ref{G2}), (\ref{G3}) into one reward function $R_A$:
\begin{equation}
R_A(s_t) = \left\{ \begin{array}{lcl}
D_t - \zeta [\mathit{Var_t}]^2 & \mbox{if} & B_t>0  \\
-\digamma & \mbox{if} & B_t=0
\end{array}\right.
\label{r_a}
\end{equation}
The agent receives its reward by maximizing the duty cycles $D_t$. 
To punish variance, we reduce the reward with the squared variance, scaled by a damping factor $\zeta$.
In the case of a failure, the reward is negative, using the punishment term $\digamma$. 
The actual values of $\digamma$ and $\zeta$ are hyperparameters of the reward function, and we will have a detailed look at them in the next section.

We generalized the reward function in (\ref{r_a}) to accommodate other RL techniques of sparse reward assignments. In these techniques, the agent gets a reward at the end of a training episode $E$. An episode consists of a set of epochs (time steps) that range from $t =1$ to $t=T$, thus $E=\{1, 2, ...,t, ..., T\}$
\begin{equation}
R_A(s_T) =\sum\limits_{t=1}^{T} \left\{ \begin{array}{lcl}
D_t - \zeta [\mathit{Var_t}]^2 & \mbox{if} & B_t>0  \\
-\digamma  & \mbox{if} & B_t=0
\end{array}\right.
\label{r_a2}
\end{equation}
If the reward is assigned at every epoch, $T=1$ and (\ref{r_a2}) reduces to (\ref{r_a}).

We considered the relevant attributes of the environment in the definition of state space. These attributes are observations of the current level in the energy buffer, current harvested energy, and the weather forecast of the whole episode. To simulate the harvested energy, we use solar radiation data to calculate the harvested energy in every hour. The weather forecast information is included to enhance performance by giving an estimate of the expected solar energy for that particular episode. This weather information can be acquired from external sources or real prediction algorithms with sufficient accuracy as the one presented in~\cite{Kraemer:2017hg}. For our case study, we simulate weather information by calculating the total harvested energy in a particular day and introduce a $\SI{20}{\percent}$ error to mimic inaccuracies in the weather forecast. We also include the previous duty cycle $D_{\mathit{t-1}}$ in the state space, so the agent makes an informed decision to avoid high variability in the duty cycle. The agent takes  the observation as input summarized in a vector of four continuous values
\begin{equation}
s_t= [B_t, \mathit{Eh_t}, \mathit{Wf_t}, D_{\mathit{t-1}}].
\end{equation}

PPO allows us to use a continuous action space corresponding to setting the operational duty cycle of a node. This enables more accurate control of the consumed energy, and therefore the utility of a node. Therefore, the action space is
\begin{equation}
a_t\in\sA= [D_{\mathit{min}}, D_{\mathit{max}}].
\end{equation}
Depending on IoT application requirements, the minimum value of the duty cycle can be set accordingly. For our case, we set it to $D_{min}=0$.

\section{Results and Discussion}

We conducted a series of simulations to systematically explore the influence of the designed reward function $R_A$ and training hyperparameters on the agent behavior. 
In total, we trained more than 300 agents on data of the year 2010 and tested their performance on data of the year 2011, using different values for the damping factor ($\zeta$) for the reward function. 
For the PPO algorithm, we also trained with different values for the hyperparameters, namely the learning rate ($\alpha$), batch size (number of state transitions used to calculate policy gradient),

discount factor ($\gamma$) and trace decay parameter ($\lambda$) of the advantage function \cite{Schulman2017}. In the following, we discuss the learning rate $\alpha$ and damping factor $\zeta$ in more detail.

\begin{figure}[tbp]
\centerline{\includegraphics[width=\linewidth]{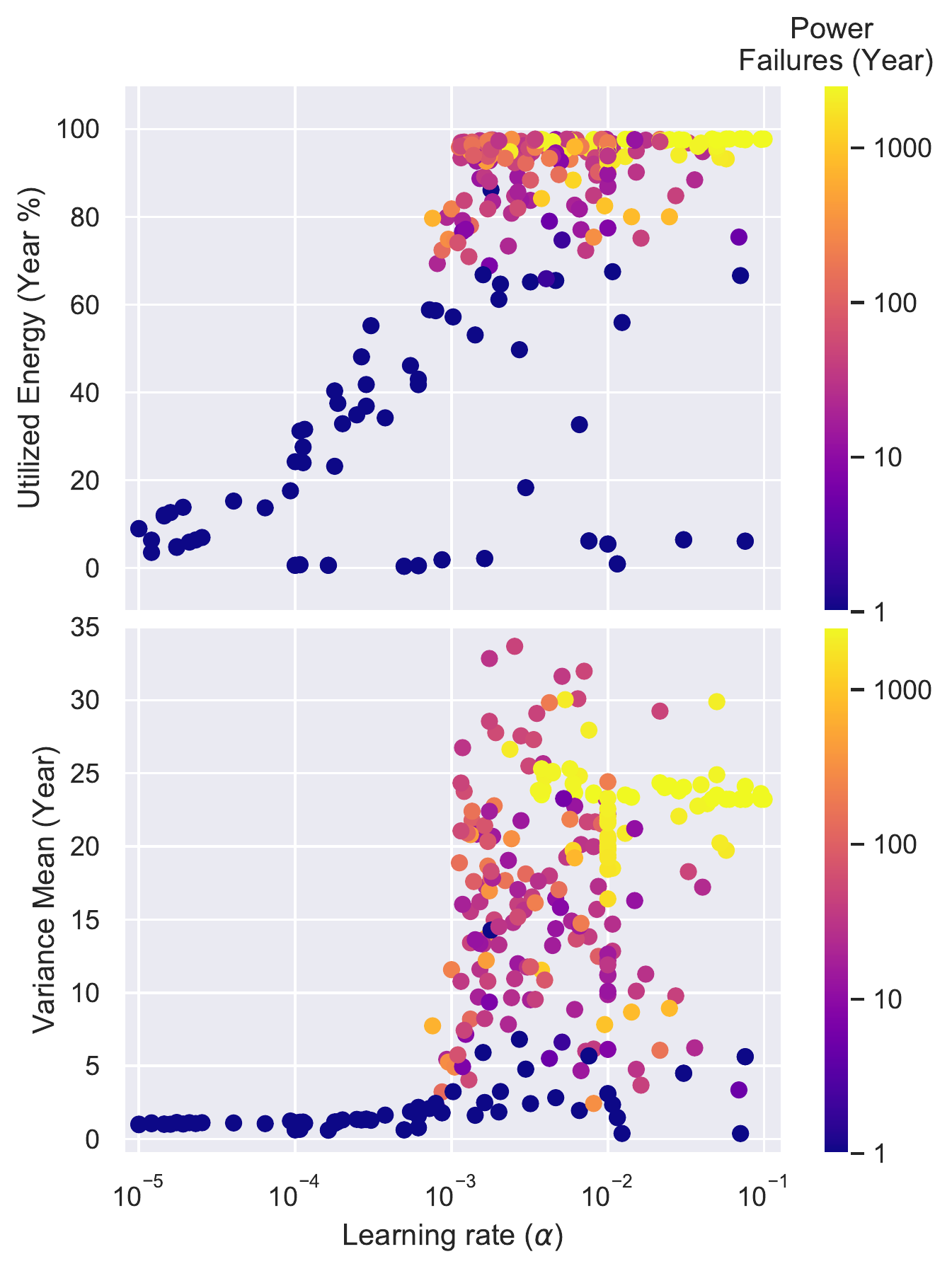}}
\caption{Influence of the learning rate ($\alpha$)  on agent's learning process in terms of variance in duty cycle and utilized energy.}
\label{tradeoff_lr}
\end{figure}

The learning rate ($\alpha$) has a significant impact on the learning process as shown in Fig.~\ref{tradeoff_lr}. The x-axis corresponds to the learning rate, which we chose to tune in the range of  $10^{-5}$ to $10^{-1}$. The y-axis in the upper plot corresponds to the yearly utilized energy (normalized by the maximum possible utilization), the y-axis in the lower plot corresponds to the mean-variance in duty cycle over the whole year. Each dot in the figure represents a single agent, and its color indicates the number of times the agent has emptied the energy buffer, i.e., failed. Training with low learning rates ($\alpha < 10^{-3}$) results in agents that learn poorly or not at all. These agents pick constant and low duty cycles, which results in low variance and few power failures, thus fulfilling application goals (\ref{G2}) and (\ref{G3}), but failing to achieve the goal of maximizing utilized energy in (\ref{G1}).

\begin{figure}[t!]
\centerline{\includegraphics[width=\linewidth]{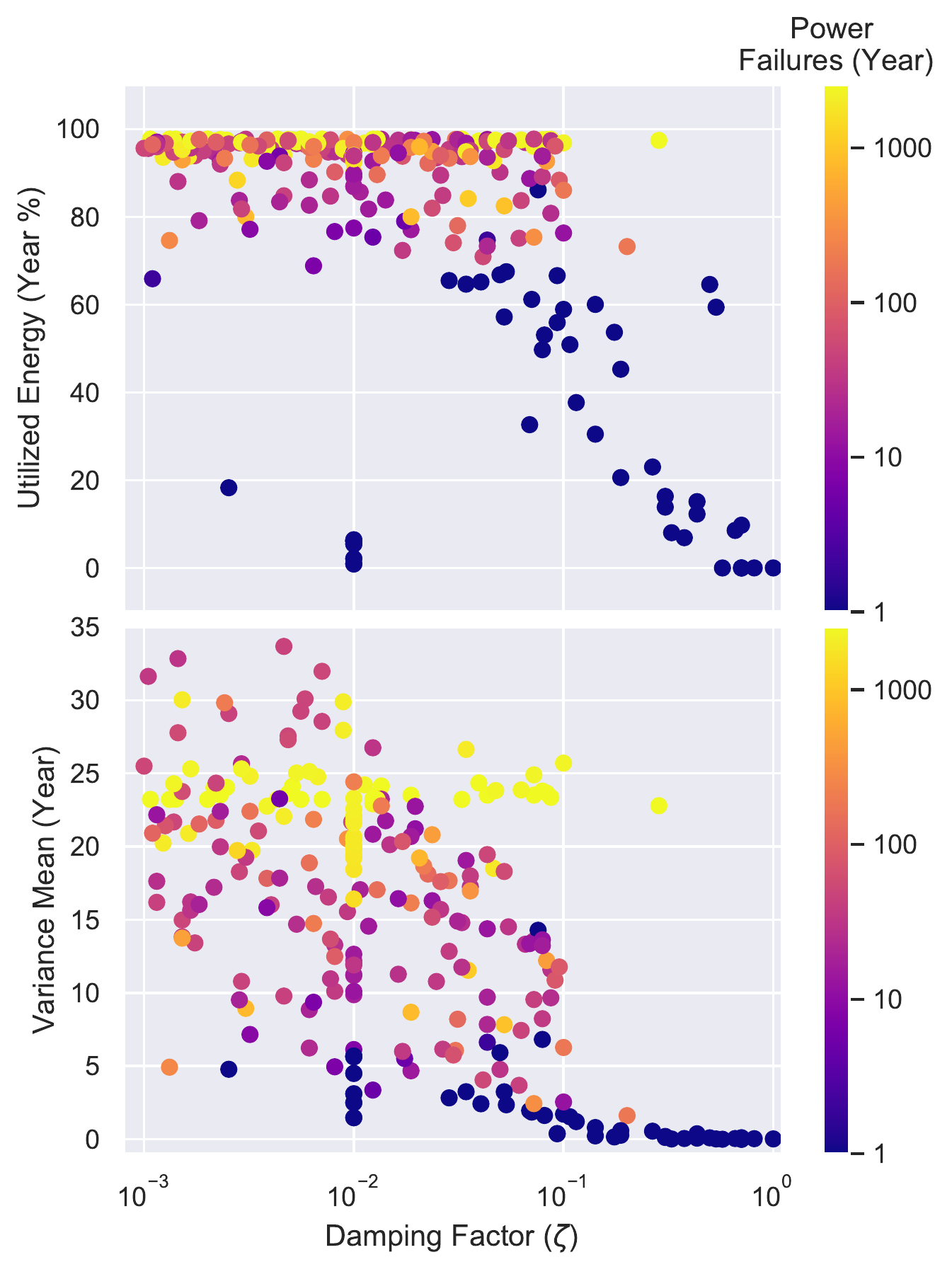}}
\caption{Influence of the damping factor ($\zeta$) in $R_A(t)$ on an IoT node's behavior in terms of utilized energy and variance in duty cycle.}
\label{tradeoff_var}
\end{figure}

The choice of damping factor $\zeta$ in the reward function also has a significant impact on the resulting behavior.
Figure~\ref{tradeoff_var} shows the results of all trained agents running over the whole year of 2011 for different damping factors $\zeta$ on the x-axis.
We observe that agents trained with values of $\zeta$ in the approximate range ($10^{-2}< \zeta < 10^{-1}$) display the desired behavior of less variability and better energy utilization, which translates to an appropriate, continuous measurement coverage of the IoT node.
We also observe that training with $\zeta$ in the approximate range ($\zeta \geq 10^{-1}$) leads to overdamped agents, which have less variance and fewer power failures, but also less energy utilization. Contrarily, most agents trained with values of $\zeta$ in the approximate range ($\zeta \geq 0.1$) are underdamped, with few exceptions as outliers. These underdamped agents have high utilized energy, but also higher variability in the duty cycle and more frequent power failures.

\begin{figure}[h!]
\centerline{\includegraphics[width=\linewidth]{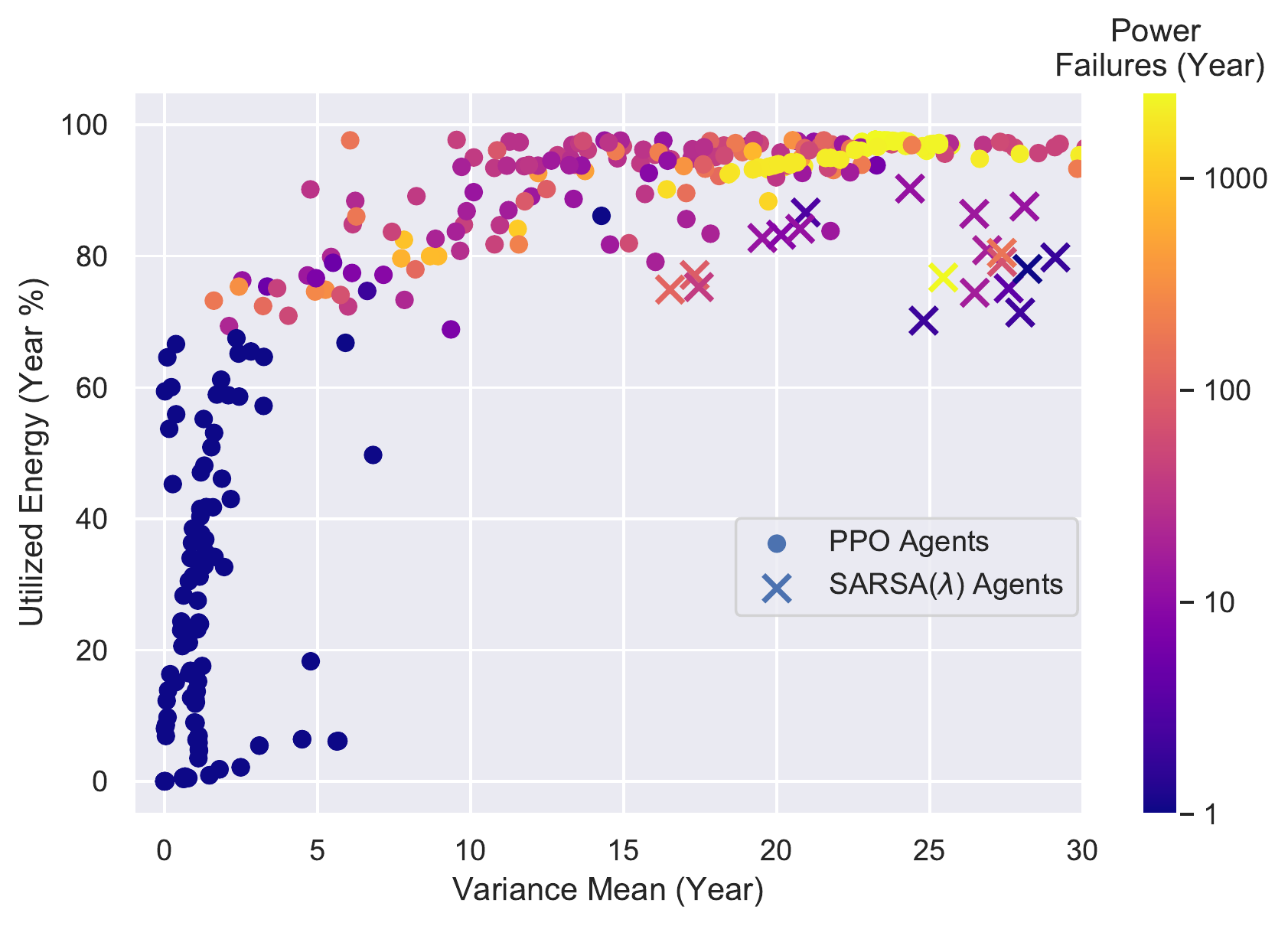}}
\caption{Performance comparison of SARSA($\lambda$) agents with PPO agents using the reward function $R_{\mathit{A}}$}
\label{sarsa_above_ppo}
\end{figure}

Depending on the choice of damping factor $\zeta$ and failure penalty $\digamma$ in the reward function $R_A$, agents try to maximize rewards by prioritizing among the conflicting  objective (\ref{G1}), (\ref{G2}), (\ref{G3}). Accordingly, agents learn various policies that achieve different performance on each objective, as shown in Fig.~\ref{sarsa_above_ppo}. This enables choosing between different trade-off agents based on the context of individual IoT applications.

In comparison, using the same reward function $R_{A}$ for SARSA-trained
agents, achieves mediocre results at best, as shown in
Fig.~\ref{sarsa_above_ppo}.  The SARSA agents (shown as crosses) score poorly
in terms of energy utilization and have a high variance.  We attribute their
relatively low failure rate to low utilization.  This is as expected since the
discretization necessary for SARSA does not offer the granularity needed for
effective learning.

\begin{figure*}[tbp]
\centerline{\includegraphics[width=\textwidth]{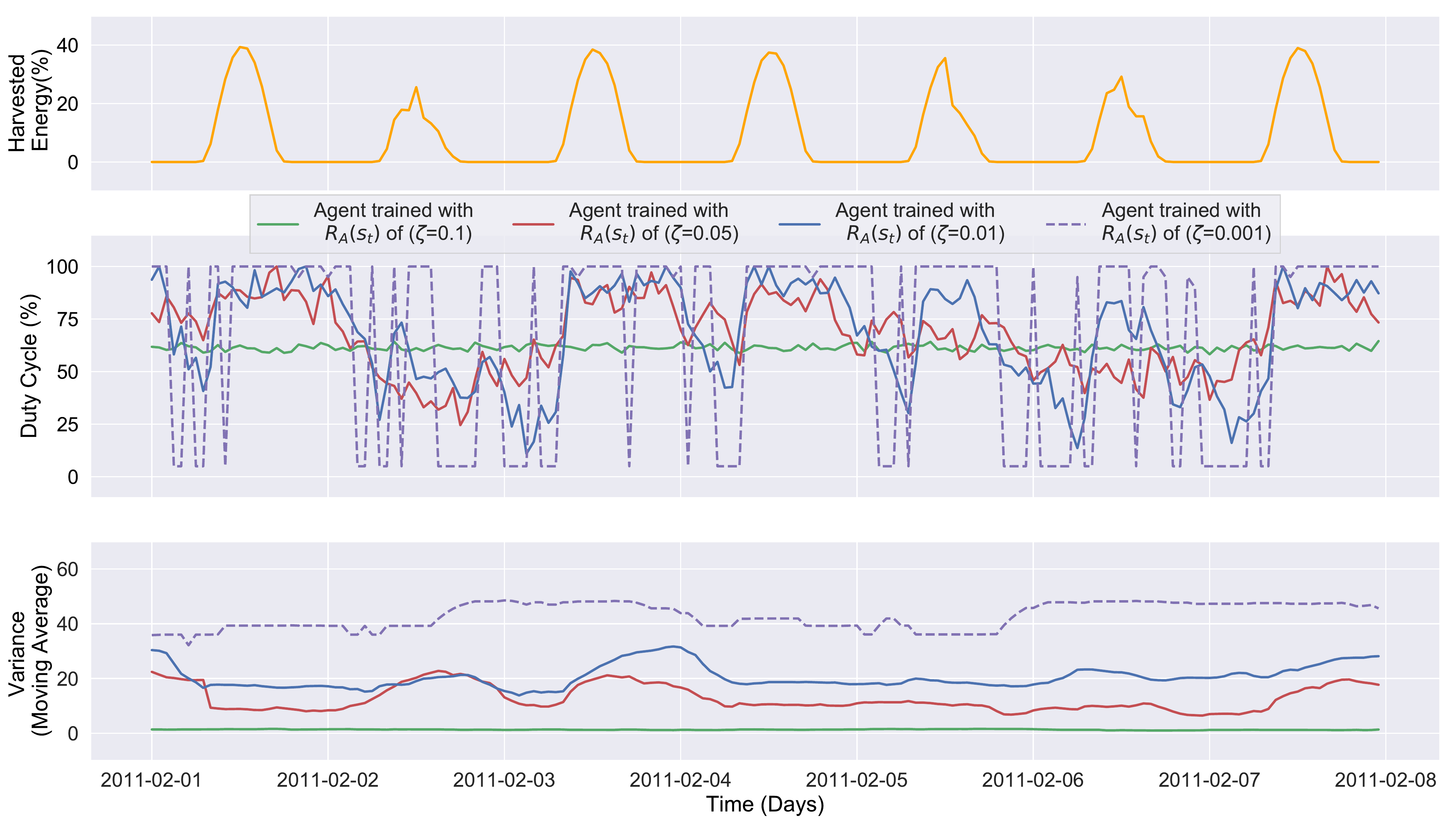}}
\caption{Performance results over one week (Tokyo 2011) of four selected agents trained with different damping factors ($\zeta$) in their reward function $R_A(t)$ }
\label{agents_feb}
\end{figure*}

We also take a more detailed look at the behavior over a one-week window. 
Figure~\ref{agents_feb} shows the performance of four selected agents trained with different damping factors $\zeta$. 
The first day yields much solar power, so all agents utilize this energy by setting their duty cycles to high values. Since the weather forecast for the next day indicates less energy, all agents reduce their duty cycles. However, the pattern of the decrease differs according to the damping factor. Efficiently damped agents show the desired gradual decrease, while underdamped agents increase sharply and oscillate between maximum and minimum duty cycles.  When more energy is available again, we see analogous behavior when the duty cycles increase. The overdamped agents appear unaffected by the variance in energy supply, but they exhibit an overall low energy utilization. For example, the agent trained with  $\zeta = 0.1$  avoids a high variance penalty by setting its duty cycle to a constant value, which leaves energy unutilized in times when much solar energy is available. Table \ref{t_1} summarizes the overall performance of the agents in Fig.~\ref{agents_feb} over the whole year.

\begin{table}[]
\caption{Summary of performance results over a year of four selected agents trained with different damping factors ($\zeta$) }
\begin{center}
\begin{tabular}{@{}cccc@{}}
\toprule
$\zeta$ & Utilized Energy  & Variance Mean & Power Failures \\ \midrule
0.1   & 77\%                          & 2.5                        & 11                     \\
0.05  & 94\%                           & 9.7                        & 14                     \\
0.01  & 95\%                           & 12                        & 17                      \\
0.001 & 98\%                           & 25.6                        & 23                    \\ \bottomrule
\end{tabular}
\end {center}
\label{t_1}
\end{table}

To investigate the effect of the neural network architecture of the policy approximation on performance, we trained with a different number of hidden layers and units. 
Our results show that shallower networks perform as well as deeper networks. We ended up using an architecture layout with two hidden layers of 64 units and a tanh activation function. The output layer has two units computing the mean and standard deviation of the Gaussian policy (\ref{gaussian_policy}). Therefore, we substantiate that an agent's policy can be approximated with a neural network that requires low computational effort and memory footprint, which makes it feasible for deployment also in resource-constrained sensor nodes.

\section{Conclusion}
Using reinforcement learning (RL) to manage constrained IoT nodes provides a path to learn optimal behavior without careful human attention.
To the best of our knowledge, this is the first application of a policy-gradient method to this problem.
We have shown that state-of-the-art policy-gradient RL methods such as PPO which use neural networks as function approximators are suitable for the use in IoT, 
that they outperform older RL approaches, 
and that they can solve more difficult problems which better describe and encourage desired behavior.
Additionally, their suitability for continuous problems removes another manual design step of discretization.
We also investigated the influence of hyperparameters on the resulting policies.
While we have focused on the maximization of duty cycle with minimal variance, 
we believe there is potential to solve much more complex problems.
The work presented hence leads to more autonomy in IoT systems, 
in which RL takes care of technicalities so that engineers can focus on the real application goals.
This will allow a broader application of IoT in complex domains.

\section*{Acknowledgments}
This research was supported by the Office of Naval Research (N0001418WX01582) and the Department of Defense High Performance Computing Modernization Program.

\balance{}

\bibliographystyle{IEEEtran}

\end{document}